\pdfoutput=1

\documentclass[11pt,a4paper]{article}
\usepackage[hyperref]{emnlp2020}
\usepackage{times}
\usepackage{latexsym}

\usepackage{microtype}

\usepackage{adjustbox}
\usepackage{url}
\usepackage{algorithm2e}
\usepackage{amssymb}
\usepackage{amsmath}
\usepackage{arydshln}
\usepackage{verbatim}
\usepackage{booktabs}
\usepackage{graphicx} %package to manage images
\graphicspath{ {./images/} }

\usepackage{booktabs}
\usepackage{xcolor}
\usepackage{multirow}
\definecolor{amber}{rgb}{1.0, 0.49, 0.0}
\definecolor{applegreen}{rgb}{0.55, 0.71, 0.0}

\aclfinalcopy % Uncomment this line for the final submission
\title{Hierarchical Graph Network for Multi-hop Question Answering}

\author{\textbf{Yuwei Fang},\hspace{1mm} \textbf{Siqi Sun},\hspace{1mm} \textbf{Zhe Gan},\hspace{1mm} \textbf{Rohit Pillai},\hspace{1mm} \textbf{Shuohang Wang},\hspace{1mm} \textbf{Jingjing Liu}\\
Microsoft Dynamics 365 AI Research\\
{\tt \small{ \{yuwfan,siqi.sun,zhe.gan,rohit.pillai,shuohang.wang,jingjl\}@microsoft.com}}}

\date{}

\begin{document}
\maketitle
\begin{abstract}
  In this paper, we present Hierarchical Graph Network (HGN) for multi-hop question answering. To aggregate clues from scattered texts across multiple paragraphs, a hierarchical graph is created by constructing nodes on different levels of granularity (questions, paragraphs, sentences, entities), the representations of which are initialized with pre-trained contextual encoders. Given this hierarchical graph, the initial node representations are updated through graph propagation, and multi-hop reasoning is performed via traversing through the graph edges for each subsequent sub-task (e.g., paragraph selection, supporting facts extraction, answer prediction). By weaving heterogeneous nodes into an integral unified graph, this hierarchical differentiation of node granularity enables HGN to support different question answering sub-tasks simultaneously. Experiments on the HotpotQA benchmark demonstrate that the proposed model achieves new state of the art, outperforming existing multi-hop QA approaches.\footnote{Code will be released at \href{https://github.com/yuwfan/HGN}{https://github.com/yuwfan/HGN}.}
\end{abstract}

%\vspace{-1mm}
\section{Introduction}

In contrast to one-hop question answering~\cite{rajpurkar2016squad,trischler2016newsqa,lai2017race} where answers can be derived from a single paragraph~\cite{matchlstm2016,bidaf,liu2017stochastic,devlin2018bert}, many recent studies on question answering focus on multi-hop reasoning across multiple documents or paragraphs. Popular tasks include WikiHop~\cite{welbl2018constructing}, ComplexWebQuestions~\cite{talmor2018web}, and HotpotQA~\cite{yang2018hotpotqa}.

\begin{figure}[t!]
\centering
{\includegraphics[width=0.98\linewidth]{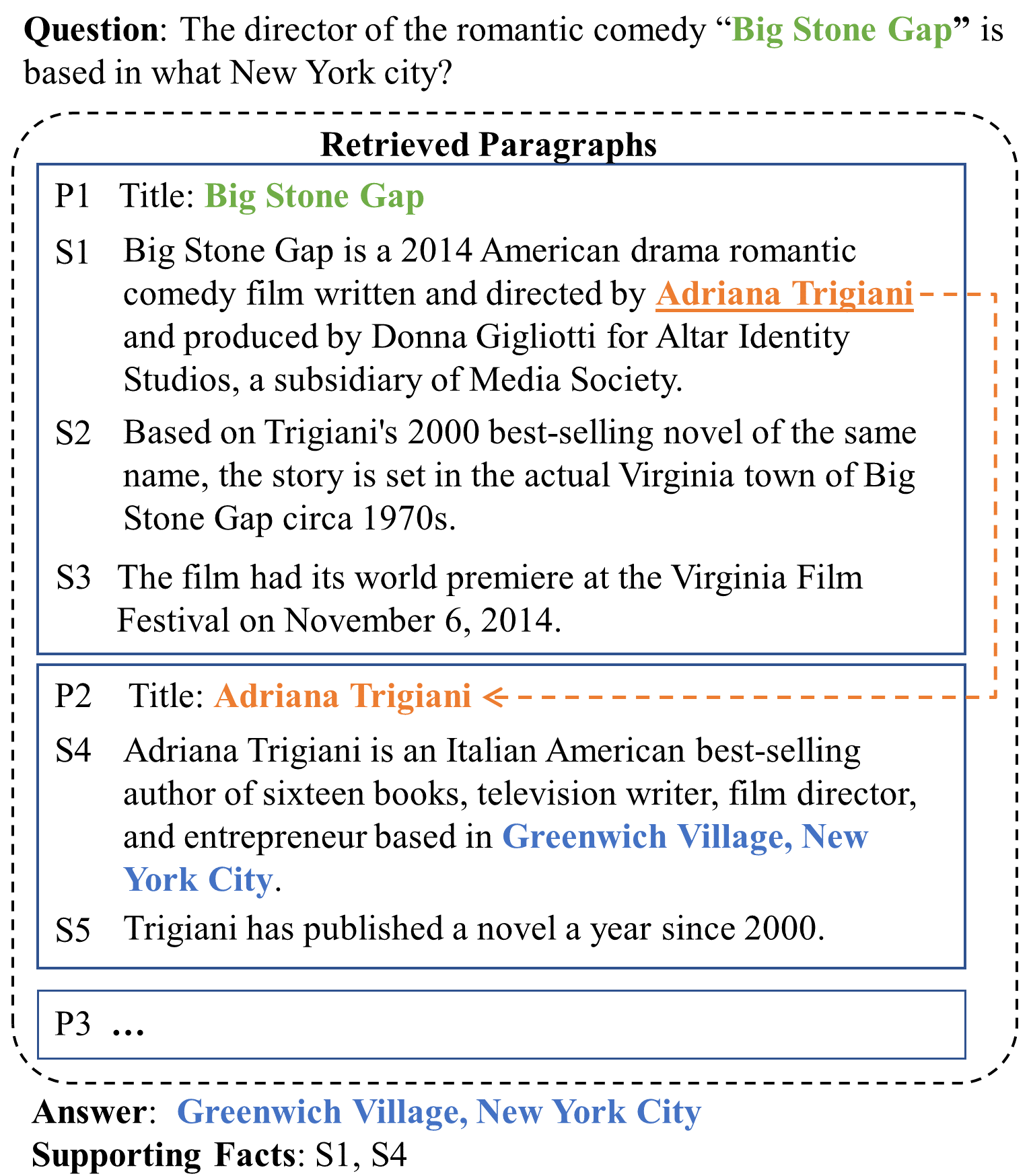}}
%\vspace{-2mm}
\caption{\label{fig:example_question} An example of multi-hop question answering from HotpotQA. The model needs to identify relevant paragraphs, determine supporting facts, and then predict the answer correctly.}
%\vspace{-5mm}
\end{figure}

An example from HotpotQA is illustrated in Figure \ref{fig:example_question}. In order to correctly answer the question (``The director of the romantic comedy `Big Stone Gap' is based in what New York city''), the model is required to first identify \textit{P1} as a relevant paragraph, whose title contains the keywords that appear in the question (``\emph{Big Stone Gap}''). \textit{S1}, the first sentence of \textit{P1}, is then chosen by the model as a supporting fact that leads to the next-hop paragraph \textit{P2}. Lastly, from \textit{P2}, the span ``\emph{Greenwich Village, New York City}'' is selected as the predicted answer.

Most existing studies use a retriever to find paragraphs that contain the right answer to the question (\textit{P1} and \textit{P2} in this case). A Machine Reading Comprehension (MRC) model is then applied to the selected paragraphs for answer prediction~\cite{nishida2019answering,min2019multi}. However, even after successfully identifying a reasoning chain through multiple paragraphs, it still remains a critical challenge how to aggregate evidence from scattered sources on different granularity levels (e.g., paragraphs, sentences, entities) for joint answer and supporting facts prediction.

To better leverage fine-grained evidences, some studies apply entity graphs through query-guided multi-hop reasoning. Depending on the characteristics of the dataset, answers can be selected either from entities in the constructed entity graph~\cite{song2018exploring, dhingra2018neural, de2018question, tu2019multi, ding2019cognitive}, or from spans in documents by fusing entity representations back into token-level document representation~\cite{DFGN}. However, the constructed graph is mostly used for answer prediction only, while insufficient for finding supporting facts. Also, reasoning through a simple entity graph~\cite{ding2019cognitive} or paragraph-entity hybrid graph~\cite{tu2019multi} lacks the ability to support complicated questions that require multi-hop reasoning.%~\cite{DFGN}. 

Intuitively, given a question that requires multiple hops through a set of documents to reach the right answer, a model needs to: ($i$) identify paragraphs relevant to the question; ($ii$) determine strong supporting evidence in those paragraphs; and ($iii$) pinpoint the right answer following the garnered evidence. To this end, Graph Neural Network with its inherent message passing mechanism that can pass on multi-hop information through graph propagation, has great potential of effectively predicting both supporting facts and answer simultaneously for complex multi-hop questions. 

\begin{figure*}[t!]
\centering
{\includegraphics[width=\linewidth]{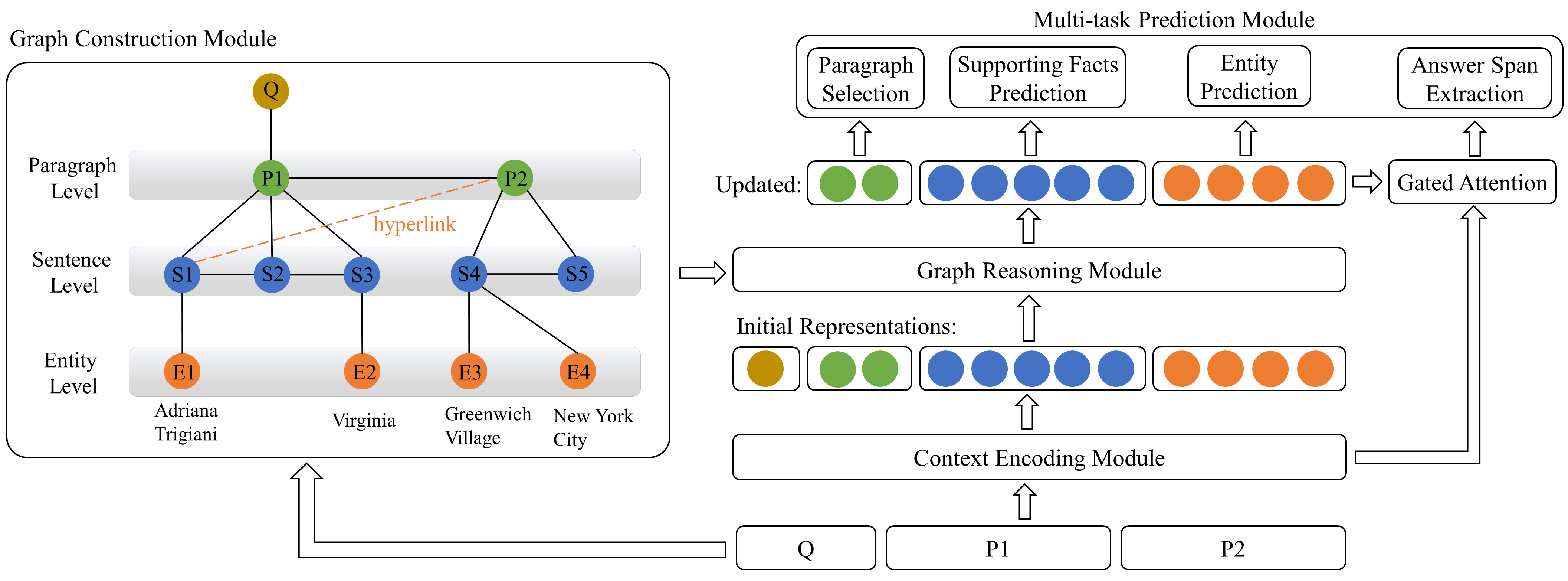}}
\caption{\label{fig:model} Model architecture of Hierarchical Graph Network. The constructed graph corresponds to the example in Figure~\ref{fig:example_question}. {\color{applegreen}Green}, {\color{blue}blue}, {\color{amber}orange}, and {\color{brown}brown} colors represent paragraph (P), sentence (S), entity (E), and question (Q) nodes, respectively. Some entities and hyperlinks are omitted for simplicity.}
%\vspace{-1mm}
\end{figure*}

Motivated by this, we propose a \textbf{H}ierarchical \textbf{G}raph \textbf{N}etwork (\textsc{HGN}) for multi-hop question answering, which empowers joint answer/evidence prediction via multi-level fine-grained graphs in a hierarchical framework. Instead of only using entities as nodes, for each question we construct a hierarchical graph to capture clues from sources with different levels of granularity. Specifically, four types of graph node are introduced: \emph{questions}, \emph{paragraphs}, \emph{sentences} and \emph{entities} (see Figure~\ref{fig:model}). To obtain contextualized representations for these hierarchical nodes, large-scale pre-trained language models such as BERT~\cite{devlin2018bert} and RoBERTa~\cite{liu2019roberta} are used for contextual encoding. These initial representations are then passed through a Graph Neural Network for graph propagation. The updated node representations are then exploited for different sub-tasks (e.g., paragraph selection, supporting facts prediction, entity prediction). Since answers may not be entities in the graph, a span prediction module is also introduced for final answer prediction.

% summarize our contribution
The main contributions of this paper are three-fold: ($i$) We propose a Hierarchical Graph Network (HGN) for multi-hop question answering, where heterogeneous nodes are woven into an integral hierarchical graph. ($ii$) Nodes from different granularity levels mutually enhance each other for different sub-tasks, providing effective supervision signals for both supporting facts extraction and answer prediction. ($iii$) On the HotpotQA benchmark, the proposed model achieves new state of the art in both Distractor and Fullwiki settings.%, and comparable to state-of-the-art results 

\section{Related Work}

\paragraph{Multi-Hop QA}
Multi-hop question answering requires a model to aggregate scattered pieces of evidence across multiple documents to predict the right answer. WikiHop \cite{welbl2018constructing} and HotpotQA \cite{yang2018hotpotqa} are two recent datasets designed for this purpose. 
Existing work on HotpotQA Distractor setting focuses on converting the multi-hop reasoning task into single-hop sub-problems. Specifically, QFE~\cite{nishida2019answering} regards evidence extraction as a query-focused summarization task, and reformulates the query in each hop. DecompRC~\cite{min2019multi} decomposes a compositional question into simpler sub-questions and leverages single-hop MRC models to answer the sub-questions. A neural modular network is also proposed in~\citet{jiang2019self}, where neural modules are dynamically assembled for more interpretable multi-hop reasoning.
Recent studies~\cite{chen2019understanding,min2019compositional,jiang2019avoiding} have also studied the multi-hop reasoning behaviors that models have learned in the task.

\paragraph{Graph Neural Network}
Recent studies on multi-hop QA also build graphs based on entities and reasoning over the constructed graph using graph neural networks~\cite{kipf2016semi,GAT}. MHQA-GRN \cite{song2018exploring} and Coref-GRN \cite{dhingra2018neural} construct an entity graph based on co-reference resolution or sliding windows. Entity-GCN \cite{de2018question} considers three different types of edges that connect different entities in the entity graph.
HDE-Graph \cite{tu2019multi} enriches information in the entity graph by adding document nodes and creating interactions among documents, entities and answer candidates.
Cognitive Graph QA \cite{ding2019cognitive} employs an MRC model to predict answer spans and possible next-hop spans, and then organizes them into a cognitive graph.
DFGN \cite{DFGN} constructs a dynamic entity graph, where in each reasoning step irrelevant entities are softly masked out and a fusion module is designed to improve the interaction between the entity graph and documents.

More recently, SAE~\cite{tu2019select} defines three types of edge in the sentence graph based on the named entities and noun phrases appearing in the question and sentences.
C2F Reader~\cite{c2f} uses graph attention or self-attention on entity graph, and argues that this graph may not be necessary for multi-hop reasoning.
\citet{asai2020learning} proposes a new graph-based recurrent method to find evidence documents as reasoning paths, which is more focused on information retrieval.
Different from the above methods, our proposed model constructs a hierarchical graph, effectively exploring relations on different granularities and employing different nodes to perform different tasks.

\paragraph{Hierarchical Coarse-to-Fine Modeling}
Previous work on hierarchical modeling for question answering is mainly based on a coarse-to-fine framework.
~\citet{choi-etal-2017-coarse} proposes to use reinforcement learning to first select relevant sentences and then produce answers from those sentences.
\citet{min-etal-2018-efficient} investigates the minimal context required to answer a question, and observes that most questions can be answered with a small set of sentences.
\citet{swayamdipta2018multimention} constructs lightweight models and combines them into a cascade structure to extract the answer.
\citet{zhong2018coarsegrain} proposes to use hierarchies of co-attention and self-attention to combine information from evidence across multiple documents. 
Different from the above methods, our proposed model organizes different granularities in a hierarchical manner and leverages graph neural network to obtain the representations for different downstream tasks.

\section{Hierarchical Graph Network}

As illustrated in Figure~\ref{fig:model}, the proposed Hierarchical Graph Network (HGN) consists of four main components: ($i$) Graph Construction Module (Sec.~\ref{sec:buildGraph}), through which a hierarchical graph is constructed to connect clues from different sources; ($ii$) Context Encoding Module (Sec.~\ref{sec:context_encoding}), where initial representations of graph nodes are obtained via a RoBERTa-based encoder; ($iii$) Graph Reasoning Module (Sec.~\ref{sec:reasoning}), where  graph-attention-based message passing algorithm is applied to jointly update node representations; and ($iv$) Multi-task Prediction Module (Sec.~\ref{sec:prediction}), where multiple sub-tasks, including paragraph selection, supporting facts prediction, entity prediction, and answer span extraction, are performed simultaneously.

\subsection{Graph Construction}
\label{sec:buildGraph}
The hierarchical graph is constructed in two steps: ($i$) identifying relevant multi-hop paragraphs; and ($ii$) adding edges representing connections between sentences/entities within the selected paragraphs.

\paragraph{Paragraph Selection}
%Starting from the question, the first step is to identify relevant paragraphs (\emph{i.e.}, the first hop). 
We first retrieve paragraphs whose titles match any phrases in the question (title matching). In addition, we train a paragraph ranker based on a pre-trained RoBERTa encoder, followed by a binary classification layer, to rank the probabilities of whether the input paragraphs contain the ground-truth supporting facts. If multiple paragraphs are found by title matching, only two paragraphs with the highest ranking scores are selected.
If title matching returns no results, we further search for paragraphs that contain entities appearing in the question. If this also fails, the paragraph ranker will select the paragraph with the highest ranking score. The number of selected paragraphs in the first-hop is at most 2.

Once the first-hop paragraphs are identified, the next step is to find facts and entities within the paragraphs that can lead to other relevant paragraphs (\emph{i.e,}, the second hop). Instead of relying on entity linking, which could be noisy, we use hyperlinks (provided by Wikipedia) in the first-hop paragraphs to discover second-hop paragraphs. Once the links are selected, we add edges between the sentences containing these links (source) and the paragraphs that the hyperlinks refer to (target), as illustrated by the dashed orange line in Figure~\ref{fig:model}. In order to allow information flow from both directions, the edges are considered as bidirectional.

Through this two-hop selection process, we are able to obtain several candidate paragraphs. In order to reduce introduced noise during inference, we use the paragraph ranker to select paragraphs with top-$N$ ranking scores in each step. 

\paragraph{Nodes and Edges}
Paragraphs are comprised of sentences, and each sentence contains multiple entities. This graph is naturally encoded in a hierarchical structure, and also motivates how we construct the hierarchical graph. For each paragraph node, we add edges between the node and all the sentences in the paragraph. For each sentence node, we extract all the entities in the sentence and add edges between the sentence node and these entity nodes. Optionally, edges between paragraphs and edges between sentences can also be included in the final graph.

Each type of these nodes captures semantics from different information sources. Thus, the hierarchical graph effectively exploits the structural information across all different granularity levels to learn fine-grained representations, which can locate supporting facts and answers more accurately than simpler graphs with homogeneous nodes. 

An example hierarchical graph is illustrated in Figure~\ref{fig:model}.  
%Black, green, purple and yellow are used to represent question, paragraph, sentence and entity nodes respectively. 
We define different types of edges as follows: ($i$) edges between question node and paragraph nodes; ($ii$) edges between question node and its corresponding entity nodes (entities appearing in the question, not shown for simplicity); ($iii$) edges between paragraph nodes and their corresponding sentence nodes (sentences within the paragraph); ($iv$) edges between sentence nodes and their linked paragraph nodes (linked through hyperlinks); ($v$) edges between sentence nodes and their corresponding entity nodes (entities appearing in the sentences); ($vi$) edges between paragraph nodes; and ($vii$) edges between sentence nodes that appear in the same paragraph. Note that a sentence is only connected to its previous and next neighboring sentence. The final graph consists of these seven types of edges as well as four types of nodes, which link the question to paragraphs, sentences, and entities in a hierarchical way.

\subsection{Context Encoding} \label{sec:context_encoding}
Given the constructed hierarchical graph, the next step is to obtain the initial representations of all the graph nodes. To this end, we first combine all the selected paragraphs into context $C$, which is concatenated with the question $Q$ and fed into pre-trained Transformer RoBERTa, followed by a bi-attention layer \cite{bidaf}. We denote the encoded question representation as $\mathbf{Q}=\{\mathbf{q}_0, \mathbf{q}_1, \ldots, \mathbf{q}_{m-1}\} \in \mathbb{R}^{m \times d}$, and the encoded context representation as  $\mathbf{C}=\{\mathbf{c}_0, \mathbf{c}_1, ..., \mathbf{c}_{n-1}\} \in \mathbb{R}^{n \times d}$, where $m$, $n$ are the length of the question and the context, respectively. Each $\mathbf{q}_i$ and $\mathbf{c}_j \in\mathbb{R}^{d}$. 

A shared BiLSTM is applied on top of the context representation $\mathbf{C}$, and the representations of different nodes are extracted from the output of the BiLSTM, denoted as $\mathbf{M} \in \mathbb{R}^{n \times 2d}$. 
For entity/sentence/paragraph nodes, which are spans of the context, the representation is calculated from: ($i$) the hidden state of the backward LSTM at the start position, and ($ii$) the hidden state of the forward LSTM at the end position. For the question node, a max-pooling layer is used to obtain its representation. Specifically,
\begin{align}
    \nonumber
    \mathbf{p}_i &= \text{MLP}_1 \left(\left[\mathbf{M}[P^{(i)}_{start}][d{:}]; \mathbf{M}[P^{(i)}_{end}][{:}d]\right] \right) \\
    \nonumber
    \mathbf{s}_i &= \text{MLP}_2\left(\left[\mathbf{M}[S^{(i)}_{start}][d{:}]; \mathbf{M}[S^{(i)}_{end}][{:}d]\right] \right) \\
    \nonumber
    \mathbf{e}_i &= \text{MLP}_3\left(\left[\mathbf{M}[E^{(i)}_{start}][d{:}]; \mathbf{M}[E^{(i)}_{end}][{:}d]\right] \right) \\
    \mathbf{q} &= \text{max-pooling}(\mathbf{Q})\,,
\end{align}
where $P^{(i)}_{start}$, $S^{(i)}_{start}$, and $E^{(i)}_{start}$  denote the start position of the $i$-th paragraph/sentence/entity node. Similarly, $P^{(i)}_{end}$, $S^{(i)}_{end}$, and $E^{(i)}_{end}$ denote the corresponding end positions. $\text{MLP}(\cdot)$ denotes an MLP layer, and $[;]$ denotes the concatenation of two vectors. As a summary, after context encoding, each $\mathbf{p}_i$, $\mathbf{s}_i$, and $\mathbf{e}_i \in \mathbb{R}^{d}$, serves as the representation of the $i$-th paragraph/sentence/entity node. The question node is represented as $\mathbf{q} \in \mathbb{R}^{d}$.

\subsection{Graph Reasoning} \label{sec:reasoning}
After context encoding, HGN performs reasoning over the hierarchical graph, where the contextualized representations of all the graph nodes are transformed into higher-level features via a graph neural network. Specifically, let $\mathbf{P} = \{\mathbf{p}_i\}_{i=1}^{n_p}$, $\mathbf{S} = \{\mathbf{s}_i\}_{i=1}^{n_s}$, and $\mathbf{E} = \{\mathbf{e}_i\}_{i=1}^{n_e}$, where $n_p$, $n_s$ and $n_e$ denote the number of paragraph/sentence/entity nodes in a graph. In experiments, we set $n_p=4$, $n_s=40$ and $n_e=60$ (padded where necessary), 
% \siqi{should those be the upper bound?} \zhe{fixed} 
and denote $\mathbf{H} = \{\mathbf{q},\mathbf{P},\mathbf{S},\mathbf{E}\} \in \mathbb{R}^{g\times d}$, where $g = n_p+ n_s + n_e + 1$, and $d$ is the feature dimension of each node. 
%For notational simplicity, let $h =  \{ Q, P, S, E \}$ represent all the nodes in the graph with $|Q| = 1, |P| = 4, |S| = 30 $ and $|E| = 50$. For simplicity, we redefine $h = \{h_1, h_2, .., h_{N}\}, h_i \in \mathbb{R}^{d}$ where $N = |P|+ |S| + |E| + |Q|$ and $d$ is the feature dimension of each node. 

For graph propagation, we use Graph Attention Network (GAT)~\cite{GAT} to perform message passing over the hierarchical graph. Specifically, GAT takes all the nodes as input, and updates node feature $\mathbf{h}_{i}^\prime$ through its neighbors $\mathcal{N}_{i}$ in the graph. Formally,
\begin{align}
\mathbf{h}_{i}^\prime = \text{LeakyRelu}\Big(\sum_{j \in \mathcal{N}_{i}}\alpha_{ij} \mathbf{h}_{j}\mathbf{W}\Big)\,,
\end{align}
%where $\parallel$ represents concatenation, K is the head number, $\alpha_{ij}^{k}$ is the coefficients from $k$-th independent attention.
where $\mathbf{h}_{j}$ is the $j^{\text{th}}$ vector from $\mathbf{H}$, $\mathbf{W}\in \mathbb{R}^{d\times d}$ is a weight matrix\footnote{Note that we omit the bias term for all the weight matrices in the paper to save space.} to be learned, and $\alpha_{ij}$ is the attention coefficients, which can be calculated by:
\begin{align}
\label{equ:coeff}
\alpha_{ij} = \frac{\exp(f([\mathbf{h}_i; \mathbf{h}_j]\mathbf{w}_{e_{ij}}))}{\sum_{k \in \mathcal{N}_{i}}\exp(f([\mathbf{h}_i; \mathbf{h}_k]\mathbf{w}_{e_{ik}}))} \,,
\end{align} 
where $\mathbf{w}_{e_{ij}}\in \mathbb{R}^{2d}$ is the weight vector corresponding to the edge type $e_{ij}$ between the $i$-th and $j$-th nodes, and $f(\cdot)$ denotes the LeakyRelu activation function. 
In a summary, after graph reasoning, we obtain $\mathbf{H}^\prime = \{ \mathbf{h}_0^\prime, \mathbf{h}_1^\prime, \ldots, \mathbf{h}_g^\prime \} \in \mathbb{R}^{g\times d}$, from which the updated representations for each type of node can be obtained, \emph{i.e.}, $\mathbf{P}^\prime \in \mathbb{R}^{n_p \times d}$, $\mathbf{S}^\prime \in \mathbb{R}^{n_s \times d}$, $\mathbf{E}^\prime \in \mathbb{R}^{n_e \times d}$, and $\mathbf{q}^\prime \in \mathbb{R}^{d}$.

\paragraph{Gated Attention} 
The graph information will further contribute to the context information for answer span extraction.
We merge the context representation $\mathbf{M}$ and the graph representation $\mathbf{H}'$ via a gated attention mechanism:  
\begin{align}
    \nonumber
    \mathbf{C} &= \text{Relu} (\mathbf{M}\mathbf{W}_m) \cdot \text{Relu} (\mathbf{H}'\mathbf{W}'_m)^{\text{T}} \\
    \nonumber
    \bar{\mathbf{H}} &= \text{Softmax}(\mathbf{C})\cdot \mathbf{H}'\\
    \mathbf{G} &= \sigma ( [\mathbf{M};\bar{\mathbf{H}}]\mathbf{W}_s) \cdot \text{Tanh} ( [\mathbf{M};\bar{\mathbf{H}}]\mathbf{W}_t),
    \label{eqn:gated}
\end{align}
where $\mathbf{W}_m\in \mathbb{R}^{2d\times 2d}, \mathbf{W}'_m\in \mathbb{R}^{2d\times 2d}, \mathbf{W}_s\in \mathbb{R}^{4d\times 4d}, \mathbf{W}_t\in \mathbb{R}^{4d\times 4d}$ are weight matrices to learn. 
%$\mathbf{C}\in \mathbb{R}^{n\times g}$. 
$\mathbf{G}\in \mathbb{R}^{n\times 4d}$ is the gated representation which will be used for answer span extraction.

\begin{table*}[t!]
\centering
\begin{adjustbox}{scale=0.95,center}
\begin{tabular}{lcccccc}
\toprule
\multirow{2}{*}{Model} & \multicolumn{2}{c}{Ans} & \multicolumn{2}{c}{Sup}  &  \multicolumn{2}{c}{Joint} \\ %\cline{2-7}
& EM & F1 & EM & F1 & EM & F1 \\ \midrule
DecompRC \cite{min2019multi} & 55.20 & 69.63 & - & - & - & - \\ 
ChainEx \cite{chen2019multi} & 61.20	& 74.11 & - & - & - & - \\
Baseline Model \cite{yang2018hotpotqa} & 45.60 & 59.02 & 20.32 & 64.49 & 10.83 & 40.16 \\
QFE \cite{nishida2019answering} & 53.86 & 68.06 & 57.75 & 84.49 & 34.63 & 59.61 \\
DFGN \cite{DFGN} & 56.31 & 69.69 & 51.50 & 81.62 & 33.62 & 59.82 \\
LQR-Net \cite{grail2020latent} & 60.20 & 73.78 & 56.21 & 84.09 & 36.56 & 63.68 \\
P-BERT$^\dagger$ & 61.18 & 74.16 & 51.38 & 82.76 & 35.42 & 63.79 \\
TAP2~\cite{glass2019span} & 64.99 & 78.59 & 55.47 & 85.57 & 39.77 & 69.12 \\
EPS+BERT$^\dagger$ & 65.79 & 79.05 & 58.50 & 86.26 & 42.47 & 70.48 \\
SAE-large \cite{tu2019select} & 66.92 & 79.62 & 61.53 & 86.86 & 45.36 & 71.45 \\
C2F Reader\cite{c2f} & 67.98 & 81.24 & 60.81 & 87.63 & 44.67 & 72.73 \\
Longformer$^{\star}$~\cite{Beltagy2020Longformer} & 68.00 & 81.25 & 63.09 & 88.34 & 45.91 & 73.16 \\
ETC-large$^{\star}$~\cite{zaheer2020big} & 68.12 & 81.18 & \textbf{63.25} & \textbf{89.09} &  46.40 & 73.62 \\
\midrule
%\cdashline{1-7}
HGN (ours) & \textbf{69.22} & \textbf{82.19} & 62.76 & 88.47 & \textbf{47.11} & \textbf{74.21} \\ \bottomrule
\end{tabular}
\end{adjustbox}
\caption{\label{tab:leaderboard_distractor}
Results on the test set of HotpotQA in the Distractor setting. HGN achieves state-of-the-art results at the time of submission (Dec. 1, 2019). ($\dagger$) and ($\star$) indicates unpublished and concurrent work. RoBERTa-large~\cite{liu2019roberta} is used for context encoding.}
%\vspace{-3mm}
\end{table*}

\subsection{Multi-task Prediction} \label{sec:prediction}
After graph reasoning, the updated node representations are used for different sub-tasks: ($i$) paragraph selection based on \emph{paragraph} nodes; ($ii$) supporting facts prediction based on \emph{sentence} nodes; and ($iii$) answer prediction based on \emph{entity} nodes and context representation $\mathbf{G}$. Since the answers may not reside in entity nodes, the loss for entity node only serves as a regularization term.

In our HGN model, all three tasks are jointly performed through multi-task learning. The final objective is defined as:
\begin{align}
\mathcal{L}_{joint} &= \mathcal{L}_{start} +\mathcal{L}_{end} + \lambda_{1} \mathcal{L}_{para} + \lambda_{2}\mathcal{L}_{sent}  \nonumber \\
      &+ \lambda_{3} \mathcal{L}_{entity} + \lambda_{4} \mathcal{L}_{type} \,,
\label{eq:joint}
\end{align}
where $\lambda_1$, $\lambda_2$, $\lambda_3$, and $\lambda_4$ are hyper-parameters, and each loss function is a cross-entropy loss, calculated over the logits (described below).
%\yuwei{probabilities -> logits}

For both paragraph selection ($\mathcal{L}_{para}$) and supporting facts prediction ($\mathcal{L}_{sent}$), we use a two-layer MLP as the binary classifier:
\begin{align}
    \mathbf{o}_{sent} = \text{MLP}_4(\mathbf{S}^{\prime}), \,\,
    \mathbf{o}_{para} = \text{MLP}_5(\mathbf{P}^{\prime}) \,,
\end{align}
where $\mathbf{o}_{sent} \in \mathbb{R}^{n_s}$ 
%\shuo{maybe add dimension here, a little bit confuse} 
represents whether a sentence is selected as supporting facts, and $\mathbf{o}_{para}\in \mathbb{R}^{n_p}$ represents whether a paragraph contains the ground-truth supporting facts. %$\sigma(\cdot)$ denotes the sigmoid function.

We treat entity prediction ($\mathcal{L}_{entity}$) as a multi-class classification problem. Candidate entities include all entities in the question and those that match the titles in the context. If the ground-truth answer does not exist among the entity nodes, the entity loss is zero. Specifically,
\begin{equation}
    \mathbf{o}_{entity} = \text{MLP}_6(\mathbf{E}^\prime)\,.
\end{equation}
The entity loss will only serve as a regularization term, and the final answer prediction will only rely on the answer span extraction module as follows.

The logits of every position being the start and end of the ground-truth span are computed by  a two-layer MLP on top of $\mathbf{G}$ in Eqn.(\ref{eqn:gated}):
\begin{align}
    %\nonumber
    %\mathbf{H}_{start} &= \text{BiLSTM}_{0}([\mathbf{M}; rep(\mathbf{o}_{sent} )])) \\
    %\nonumber
    %\mathbf{H}_{end} &=  \text{BiLSTM}_{1}([\mathbf{M}; \mathbf{H}_{start}; rep(\mathbf{o}_{sent})]) \\
    %\nonumber
    \mathbf{o}_{start} = \text{MLP}_7(\mathbf{G}),\,\,
    \mathbf{o}_{end} = \text{MLP}_8(\mathbf{G}) \,.
\end{align}
Following previous work~\cite{DFGN}, we also need to identify the answer type, %For answer-type\footnote{Following previous work, answer type 
which includes the types of span, entity, yes and no.
We use a 3-way two-layer MLP for answer-type classification based on the first hidden representation of $\mathbf{G}$:
\begin{align}
    %\nonumber
    %\mathbf{H}_{type} &= \text{BiLSTM}_2([\mathbf{M};  \mathbf{H}_{end}; rep(\mathbf{o}_{sent})])\\
    \mathbf{o}_{type} &= \text{MLP}_9(\mathbf{G}[0])\,.
\end{align}
During decoding, we first use this to determine the answer type. If it is ``yes'' or ``no'', we directly return it as the answer. 
Overall, the final cross-entropy loss ($\mathcal{L}_{joint}$) used for training is defined over all the aforementioned logits: $\mathbf{o}_{sent}, \mathbf{o}_{para}, \mathbf{o}_{entity}, \mathbf{o}_{start}, \mathbf{o}_{end}, \mathbf{o}_{type}$. 

\section{Experiments}
In this section, we describe experiments comparing HGN with state-of-the-art approaches and provide detailed analysis on the model and results.
\begin{table*}[t!]
\centering
\begin{adjustbox}{scale=0.95,center}
\begin{tabular}{lcccccc}
\toprule
\multirow{2}{*}{Model} & \multicolumn{2}{c}{Ans} & \multicolumn{2}{c}{Sup}  & \multicolumn{2}{c}{Joint} \\ %\cline{2-7}
& EM & F1 & EM & F1 & EM & F1 \\ \midrule
TPReasoner \cite{xiong2019simple} & 36.04 & 47.43 & - & - & - & - \\ 
Baseline Model  \cite{yang2018hotpotqa}  & 23.95 & 32.89 & 3.86 & 37.71 & 1.85 & 16.15 \\ 
QFE \cite{nishida2019answering} & 28.66 & 38.06 & 14.20 & 44.35 & 8.69 & 23.10 \\
MUPPET \cite{feldman2019multi} & 30.61 & 40.26 & 16.65 & 47.33 & 10.85 & 27.01 \\
Cognitive Graph \cite{ding2019cognitive} & 37.12	& 48.87 & 22.82 & 57.69 & 12.42 & 34.92 \\
PR-BERT$^\dagger$ & 43.33	& 53.79	& 21.90 & 59.63 & 14.50 & 39.11 \\
Golden Retriever \cite{qi2019answering} & 37.92 & 48.58 & 30.69 & 64.24 & 18.04 & 39.13 \\
Entity-centric BERT \cite{godbole2019multi} & 41.82 & 53.09 & 26.26 & 57.29 & 17.01 & 39.18 \\
SemanticRetrievalMRS \cite{nie2019revealing} & 45.32 & 57.34 & 38.67 & 70.83 & 25.14 & 47.60 \\
Transformer-XH \cite{Zhao2020Transformer-XH:} & 48.95	& 60.75	& 41.66	& 70.01 & 27.13	& 49.57 \\ %\cdashline{1-7} 
MIR+EPS+BERT$^\dagger$ & 52.86 & 64.79 & 42.75 & 72.00 & 31.19 & 54.75 \\
Graph Recur. Retriever \cite{asai2020learning} & \textbf{60.04}	& \textbf{72.96}	& 49.08	& 76.41	& 35.35	& 61.18 \\
\midrule
HGN (RoBERTa-large) & 57.85 & 69.93 & 51.01 & 76.82 & 37.17 & 60.74 \\ 
HGN (ALBERT-xxlarge-v2) & 59.74	& 71.41 & \textbf{51.03} & \textbf{77.37} & \textbf{37.92} & \textbf{62.26} \\
\bottomrule
\end{tabular}
\end{adjustbox}
\caption{\label{tab:leaderboard_fullwiki}
Results on the test set of HotpotQA in the Fullwiki setting. HGN achieves state-of-the-art results at the time of submission (Feb. 11, 2020). ($\dagger$) indicates unpublished work. RoBERTa-large~\cite{liu2019roberta} and ALBERT-xxlarge-v2~\cite{Lan2020ALBERT:} are used for context encoding, and SemanticRetrievalMRS is used for retrieval. Leaderboard: \href{https://hotpotqa.github.io/}{https://hotpotqa.github.io/}.}
%\vspace{-3mm}
\end{table*}
\subsection{Dataset}
%
%\paragraph{Dataset}
We use HotpotQA dataset~\cite{yang2018hotpotqa} for evaluation, a popular benchmark for multi-hop QA. Specifically, two sub-tasks are included in this dataset: ($i$) Answer prediction; and ($ii$) Supporting facts prediction. For each sub-task, exact match (EM) and partial match (F1) are used to evaluate model performance, and a joint EM and F1 score is used to measure the final performance, which encourages the model to take both answer and evidence prediction into consideration. 

There are two settings in HotpotQA: \emph{Distractor} and \emph{Fullwiki} setting. In the Distractor setting, for each question, two gold paragraphs with ground-truth answers and supporting facts are provided, along with 8 `distractor' paragraphs that were collected via a bi-gram TF-IDF retriever~\cite{chen2017reading}. The Fullwiki setting is more challenging, which contains the same training questions as in the Distractor setting, but does not provide relevant paragraphs for test set. To obtain the right answer and supporting facts, the entire Wikipedia can be used to find relevant documents. Implementation details can be found in Appendix~\ref{appendix:implementation}.

\subsection{Experimental Results}
%\yuwei{Do we need to include experiments with different loss in camery-ready?}\siqi{maybe add it to appendix?}
\paragraph{Results on Test Set}
% 1. Outperform all in both settings
% 2. Few of them work on both settings, both tasks
% 3. Compare with DFGN (public distractor setting)
% 4. Compare with Semantic MRS (public fullwiki setting)
Table~\ref{tab:leaderboard_distractor} and~\ref{tab:leaderboard_fullwiki} summarize results on the hidden test set of HotpotQA. In Distractor setting, HGN outperforms both published and unpublished work on every metric by a significant margin, achieving a Joint EM/F1 score of 47.11/74.21 with an absolute improvement of 2.44/1.48 over previous state of the art. In Fullwiki setting, HGN achieves state-of-the-art results on Joint EM/F1 with 2.57/1.08 improvement, despite using an inferior retriever; when using the same retriever as in SemanticRetrievalMRS~\cite{nie2019revealing}, our method outperforms by a significant margin, demonstrating the effectiveness of our multi-hop reasoning approach. In the following sub-sections, we provide a detailed analysis on the sources of performance gain on the dev set. Additional ablation study on paragraph selection is provided in Appendix~\ref{appendix:para_sel}.

\begin{table*}[t!]
\centering
\begin{adjustbox}{scale=0.90,center}
\begin{tabular}{p{2cm}p{5cm}p{3cm}p{3cm}p{1.5cm}}
\toprule
Category & Question & Answer & Prediction & Pct (\%)\\ 
\midrule
Annotation &  Were the films Tonka and 101 Dalmatians released in the same decade? & 1958 Walt Disney Western adventure film & No & 9 \\ 
\midrule
Multiple Answers & Michael J. Hunter replaced the lawyer who became the administrator of which agency? & EPA & Environmental Protection Agency & 24 \\
\midrule
Discrete Reasoning  &  Between two bands, Mastodon and Hole, which one has more members? &  Mastodon & Hole & 15 \\ \midrule
Commonsense \& External Knowledge & What is the name of second extended play by the artists of the mini-album Code\#01? & Code\#02 Pretty~ Pretty & Code\#01 Bad Girl & 16 \\
\midrule
Multi-hop & Who directed the film based on the rock opera 5:15 appeared in? & Franc Roddam & Ken Russell & 16 \\
\midrule
MRC & How was Ada Lovelace, the first computer programmer, related to Lord Byron in Childe Byron? & his daughter & strained relationship & 20 \\
\bottomrule
\end{tabular}
\end{adjustbox}
\caption{\label{table:error_example} Error analysis of HGN model. For `Multi-hop' errors, the model jumps to the wrong film (``Tommy (1975 film)'') instead of the correct one (``Quadrophenia (film)'') from the starting entity ``rock opera 5:15''. The supporting fact for the `MRC' example is ``\textit{Childe Byron is a 1977 play by Romulus Linney about the \textbf{strained relationship} between the poet, Lord Byron, and \textbf{his daughter}, Ada Lovelace}''.  }
%\vspace{-2mm}
\end{table*}

\paragraph{Effectiveness of Hierarchical Graph}
\begin{table}[t!]
%\small
\centering
\begin{adjustbox}{scale=0.95,center}
\begin{tabular}{lcccccc}
\toprule
Model & Ans F1 & Sup F1  & Joint F1 \\ \midrule
w/o Graph &  80.58 & 85.83 & 71.02 \\ 
PS Graph  & 81.68  & 88.44  & 73.83 \\
PSE Graph  & 82.10 & 88.40 & 74.13 \\
Hier. Graph  & \textbf{82.22}  & \textbf{88.58} & \textbf{74.37} \\ \bottomrule
%w/o Graph &  72.67 & 84.86 & 64.24 \\ 
%PS Graph  & 73.43  & 85.54  & 65.29 \\
%PSE Graph  & 73.34 & 85.79 & 65.47 \\
%Hier. Graph  & 74.07  & 85.62 & 66.01 \\ \hline
\end{tabular}
\end{adjustbox}
\caption{\label{table:abl_graph}Ablation study on the effectiveness of the hierarchical graph on the dev set in the Distractor setting. RoBERTa-large is used for context encoding.
}
%\vspace{-2mm}
\end{table}
% \zhe{results will be updated.}\yuwei{results have been updated with RoBERTa results.}

As described in Section \ref{sec:buildGraph}, we construct our graph with four types of nodes and seven types of edges. For ablation study, we build the graph step by step. First, we only consider edges from question to paragraphs, and from paragraphs to sentences, \emph{i.e.}, only edge type ($i$), ($iii$) and ($iv$) are considered. We call this the PS Graph. Based on this, entity nodes and edges related to each entity node (corresponding to edge type ($ii$) and ($v$)) are added. We call this the PSE Graph. Lastly, edge types ($vi$) and ($vii$) are added, resulting in the final hierarchical graph. 
%We call it a hierarchical graph as it is not a simple tree structure but follows a hierarchy. 

As shown in Table \ref{table:abl_graph}, the use of PS Graph improves the joint F1 score over the plain RoBERTa model by $2.81$ points. By further adding entity nodes, the Joint F1 increases by $0.30$ points. This indicates that the addition of entity nodes is helpful, but may also bring in noise, thus only leading to limited performance improvement. By including edges among sentences and paragraphs, our final hierarchical graph provides an additional improvement of $0.24$ points. We hypothesize that this is due to the explicit connection between sentences that leads to better representations.
%\vspace{-2mm}
\paragraph{Effectiveness of Pre-trained Language Model}
To verify the effects of pre-trained language models, we 
compare HGN with prior state-of-the-art methods using the same pre-trained language models. Results in Table \ref{table:HGN-LM} show that our HGN variants outperform DFGN, EPS and SAE, indicating the performance gain comes from better model design.
% anaylysis on reasoning types
% analysis on entity loss and entity answers

\begin{table}[t!]
\small
\centering
\begin{adjustbox}{scale=0.95,center}
\begin{tabular}{lccc}
\toprule
Model & Ans F1 & Sup F1  & Joint F1 \\ \midrule
DFGN (BERT-base) & 69.38 & 82.23 & 59.89 \\ 
EPS (BERT-wwm)$^\dagger$ & 79.05 & 86.26 & 70.48 \\
SAE (RoBERTa) & 80.75 & 87.38 & 72.75 \\
\midrule
HGN (BERT-base) & 74.76 & 86.61 & 66.90 \\
HGN (BERT-wwm) & 80.51 & 88.14 & 72.77 \\
HGN (RoBERTa) &  82.22 & 88.58 & 74.37 \\ 
HGN (ALBERT-xxlarge-v2) & \textbf{83.46} & \textbf{89.2} & \textbf{75.79} \\
\bottomrule
\end{tabular}
\end{adjustbox}
\caption{\label{table:HGN-LM}Results with different pre-trained language models on the dev set in the Distractor setting. ($\dagger$) is unpublished work with results on the test set, using BERT whole word masking (wwm).}
%\vspace{-3mm}
\end{table}

\subsection{Analysis}

\begin{table}[t!]
\centering
\begin{adjustbox}{scale=0.95,center}
\begin{tabular}{lcccc}
\toprule
Question & Ans F1 & Sup F1 & Joint F1 & Pct (\%)\\ \midrule
comp-yn & 93.45 & 94.22 & 88.50 & 6.19 \\
comp-span & 79.06 & 91.72 & 74.17 & 13.90 \\
bridge & 81.90 & 87.60 & 73.31 & 79.91\\ \bottomrule
\end{tabular}
\end{adjustbox}
\caption{\label{table:res_reasoning} Results of HGN for different reasoning types. `Pct' is short for `Percentage'.}
%\vspace{-3mm}
\end{table}

In this section, we provide an in-depth error analysis on the proposed model. HotpotQA provides two reasoning types: ``bridge'' and ``comparison''. ``Bridge'' questions require the identification of a bridge entity that leads to the answer, while ``comparison'' questions compare two entities to infer the answer, which could be \emph{yes}, \emph{no} or \emph{a span of text}. For analysis, we further split ``comparison'' questions into ``comp-yn'' and ``comp-span''. Table \ref{table:res_reasoning} indicates that ``comp-yn'' questions are the easiest, on which our model achieves 88.5 joint F1 score. HGN performs similarly on ``bridge'' and ``comp-span'' with 74 joint F1 score, indicating that there is still room for further improvement. % on the multi-hop reasoning part of our model.

To provide a more in-depth understanding of our model's weaknesses (and provide insights for future work), we randomly sample 100 examples in the dev set with the answer F1 as 0. After carefully analyzing each example, we observe that these errors can be roughly grouped into six categories:
%\siqi{we will choose one way to present examples (1) like below (2) table \ref{table:error_example} (3) appendix}
%\yuwei{we may also need to include context.}
%\siqi{contexts are too long, we may just add necessary context, like supporting fact}
($i$) \emph{Annotation}: the annotation provided in the dataset is not correct; ($ii$) \emph{Multiple Answers}: \,
questions may have multiple correct answers, but only one answer is provided in the dataset; ($iii)$ \emph{Discrete Reasoning}: this type of error often appears in ``comparison'' questions, where discrete reasoning is required to answer the question correctly; ($iv$) \emph{Commonsense \& External Knowledge}: to answer this type of question, commonsense or external knowledge is required; ($v$) \emph{Multi-hop}: the model fails to perform multi-hop reasoning, and finds the final answer from wrong paragraphs; ($vi$) \emph{MRC}: model correctly finds the supporting paragraphs and sentences, but predicts the wrong answer span.

Note that these error types are not mutually exclusive, but we aim to classify each example into only one type, in the order presented above. For example, if an error is classified as `Commonsense \& External Knowledge' type, it cannot be classified as `Multi-hop' or `MRC' error. 
Table \ref{table:error_example} shows examples from each category (the corresponding paragraphs are omitted due to space limit).  

%Table \ref{table:error_example} shows the percentage of each error type with an example. 
We observed that a lot of errors are due to the fact that some questions have multiple answers with the same meaning, such as ``\textit{a body of water} vs. \textit{creek}'', ``\textit{EPA} vs. \textit{Environmental Protection Agency}'', and ``\textit{American-born} vs. \textit{U.S. born}''. In these examples, the former is the ground-truth answer, and the latter is our model's prediction. Secondly, for questions that require commonsense or discrete reasoning (\emph{e.g.}, ``\textit{second}'' means ``\textit{Code\#02}''\footnote{Please refer to Row 4 in Table \ref{table:error_example} for more context.}, ``\textit{which band has more members}'', or ``\textit{who was born earlier}''), our model just randomly picks an entity as answer, as it is incapable of performing this type of reasoning. The majority of the errors are from either multi-hop reasoning or MRC model's span selection, which indicates that there is still room for further improvement. Additional examples are provided in Appendix~\ref{appendix:additional_examples}.

% Potential Bias towards HotpotQA
\subsection{Generalizability Discussion}
The hierarchical graph can be applied to different multi-hop QA datasets, though in this paper mainly tailored for HotpotQA. 
Here we use Wikipedia hyperlinks to connect sentences and paragraphs. An alternative way is to use an entity linking system to make it more generalizable. For each sentence node, if its entities exist in a paragraph, an edge can be added to connect the sentence and paragraph nodes. %\JJ{What does this mean? Edge connecting what nodes (sentence or paragraph)? And whose entities exist in it? (and what is 'it', sentence or paragraph?)} \yuwei{Edges connect sentence node with paragraph node if the entities within the sentence node exist in the paragraph node.}
In our experiments, we restrict the number of multi-hops to two for the HotpotQA task, which can be increased to accommodate other datasets.
The maximum number of paragraphs is set to four for HotpotQA, as we observe that using more documents within a maximum sequence length does not help much (see Table~\ref{table:MRC-PS} in the Appendix). 
To generalize to other datasets that need to consume longer documents, we can either: ($i$) use sliding-window-based method to chunk a long sequence into short ones; or ($ii$) replace the BERT-based backbone with other transformer-based models that are capable of dealing with long sequences~\cite{Beltagy2020Longformer,zaheer2020big,wang2020cluster}.

\section{Conclusion}

In this paper, we propose a new approach, Hierarchical Graph Network (HGN), for multi-hop question answering. To capture clues from different granularity levels, our HGN model weaves heterogeneous nodes into a single unified graph. Experiments with detailed analysis demonstrate the effectiveness of our proposed model, which achieves state-of-the-art performances on the HotpotQA benchmark. Currently, in the Fullwiki setting, an off-the-shelf paragraph retriever is adopted for selecting relevant context from large corpus of text. Future work includes investigating the interaction and joint training between HGN and paragraph retriever for performance improvement.   

%\clearpage
\bibliography{emnlp2020}
\bibliographystyle{acl_natbib}

\appendix
\section{Datasets}
There are two benchmark settings in HotpotQA: \emph{Distractor} and \emph{Fullwiki} setting.
They both have 90k training samples and 7.4k development samples.
In the Distractor setting, there are 2 gold paragraphs and 8 distractors.
However, 2 gold paragraphs may not be available in the Fullwiki Setting.
Therefore, the Fullwiki setting is more challenge which requires to search the entire Wikipedia to find relevant documents.
For both settings, there are 90K hidden test samples. More details about the dataset can be found in~\citet{yang2018hotpotqa}.

\section{Implementation Details}
\label{appendix:implementation}
Our implementation is based on the Transformer library \cite{wolf2019transformers}.
%, and we use %BERT-wwm (whole word masking) or  RoBERTa~\cite{liu2019roberta} for context encoding. 
To construct the proposed hierarchical graph, we use spacy\footnote{\href{https://spacy.io}{https://spacy.io}} to extract entities from both questions and sentences. The numbers of entities, sentences and paragraphs in one graph are limited to 60, 40 and 4, respectively. Since HotpotQA only requires two-hop reasoning, up to two paragraphs are connected to each question. 
%The second hop does not happen unless there is only one paragraph in 1-hop. 
Our paragraph ranking model is a binary classifier based on the RoBERTa-large model. For the Fullwiki setting, we leverage the retrieved paragraphs and the paragraph ranker provided by \citet{nie2019revealing}.
We finetune on the training set for 8 epochs, with batch size as 8, learning rate as 1e-5, $\lambda_{1}$ as 1, $\lambda_{2}$ as 5,  $\lambda_{3}$ as 1, $\lambda_4$ as 1, LSTM dropout rate as 0.3 and  GNN dropout rate as 0.3.
We search hyperparameters for learning rate from \{1e-5, 2e-5, 3e-5\} , $\lambda_{2}$ from \{1, 3, 5\} and dropout rate from \{0.1, 0.3, 0.5\}.

\section{Computing Resources}
We conduct experiments on 4 Quadro RTX 8000 GPUs. The parameters of each component in HGN are summarized in Table~\ref{table:resources}. The computation bottleneck is mainly from RoBERTa. The best model of HGN took around 12 hours for training, which is almost the same as the RoBERTa-large baseline. 
\begin{table}[h!]
%\small
\centering
% no scale because this table is next to table 2
\begin{adjustbox}{scale=0.95,center}
\begin{tabular}{lc}
\hline
Components & \#Parameters \\ \hline
RoBERTa & 355M  \\
Bi-Attention & 0.62M \\
BiLSTM & 1.44M \\
GNN & 29M \\
Multi-task Layer & 0.55M \\ \hline
\end{tabular}
\end{adjustbox}
\caption{\label{table:resources} Number of parameters for each component in HGN.}
%\vspace{-3mm}
\end{table}

%\section{Details of Building Hierarchical Graph}
%\label{appendix:preprocess}
%\subsection{Paragraph Ranker}
%Our paragraph ranking model is a binary classifier based on the BERT-base model. The input of the ranker is the question with an associated paragraph, and the output is a probability between 0 and 1. Paragraphs with gold supporting facts are treated as positive samples at training time.
%\subsection{Training parameters}
%Dropout rate of bi-attention is set to 0.2. The hidden size of BiLSTM is set to 768 and 1024 for the BERT-base and BERT-large whole word masking models, respectively. Dropout rate of BiLSTM/GAT is set to 0.3/0.5. $\lambda_{e}, \lambda_{s}, \lambda_{p}$ and $\lambda_t$ are set to 1, 5, 1 and 1.

\section{Effectiveness of Paragraph Selection} \label{appendix:para_sel}
%To further evaluate the contribution of our HGN, we perform analysis of the hierarchical graph on the development set with BERT-base model. 

% \footnote{We finetuned BERT with the released code from DFGN.}

The proposed HGN relies on effective paragraph selection to find relevant multi-hop paragraphs. 
Table \ref{table:PS} shows the performance of paragraph selection on the dev set of HotpotQA.
In DFGN, paragraphs are selected based on a threshold to maintain high recall (98.27\%), leading to a low precision (60.28\%). Compared to both threshold-based and pure Top-$N$-based paragraph selection, our two-step paragraph selection process is more accurate, achieving 94.53\% precision and 94.53\% recall. Besides these two top-ranked paragraphs, we also include two other paragraphs with the next highest ranking scores, to obtain a higher coverage on potential answers. 
%while sacrificing slightly the precision score. 
%The precision and recall of answer prediction based on these four paragraphs are slightly higher than those from the top-N-based ones.
%
Table \ref{table:MRC-PS} summarizes the results on the dev set in the Distractor setting, using our paragraph selection approach for both DFGN and the plain BERT-base model. Note that the original DFGN does not finetune BERT, leading to much worse performance. In order to provide a fair comparison, we modify their released code to allow finetuning of BERT. 
Results show that our paragraph selection method outperforms the threshold-based one in both models. 

\begin{table}[t!]
%\small
\centering
% no scale because this table is next to table 2
\begin{adjustbox}{scale=0.95,center}
\begin{tabular}{lccc}
\hline
Method & Precision & Recall & \#Para.\\ \hline
%Top 2 from ranker & 95.27 & 95.27 \\
%Top 4 from ranker & 49.72 & 99.13 \\ \hline
%2 paragraphs (ours)  & 95.13 & 95.13 \\
%4 paragraphs (ours) & 49.81 & 99.31 \\ \hline
Threshold-based & 60.28 & 98.27 & 3.26 \\
Top 2 from ranker & 93.43 & 93.43 & 2 \\ 
Top 4 from ranker & 49.39 & 98.48 & 4 \\\hline
1st hop & 96.10 & 59.74 & 1.24 \\
2 paragraphs (ours) & \textbf{94.53} & 94.53 & 2 \\
4 paragraphs (ours) & 49.45 & \textbf{98.74} & 4 \\ \hline
\end{tabular}
\end{adjustbox}
\caption{\label{table:PS}Performance of paragraph selection on the dev set of HotpotQA based on BERT-base.}
%\vspace{-3mm}
\end{table}

\begin{table}[h!]
%\small
\centering
\begin{adjustbox}{scale=0.95,center}
\begin{tabular}{lccc}
\hline
Model & Ans F1 & Sup F1  & Joint F1 \\ \hline
DFGN (paper)  & 69.38  & 82.23  & 59.89 \\ \hline
DFGN   & 
   &   & \\
+ threshold-based  & 
 71.90  & 83.57  & 63.04 \\
+ 2 para. (ours)  & 72.53  & 83.57  & 63.87 \\ 
+ 4 para. (ours)   & \textbf{72.67}  & 83.34  & 63.63 \\ \hline
BERT-base   &   &   &  \\
+ threshold-based  & 71.95  & 82.79  & 62.43 \\
+ 2 para. (ours)  & 72.42  & 83.64 & 63.94 \\
+ 4 para. (ours)  & \textbf{72.67} & \textbf{84.86}  & \textbf{64.24} \\ \hline
\end{tabular}
\end{adjustbox}
\caption{\label{table:MRC-PS}Results with selected paragraphs on the dev set in the Distractor setting. 
%\yuwei{BERT-base}
}
\end{table}

%\begin{table*}[t!]
%\small
%\centering
%\begin{tabular}{lcccccc}
%\hline
%\multirow{2}{*}{Model} & \multicolumn{2}{c}{Ans} & \multicolumn{2}{c}{Sup}  & \multicolumn{2}{c}{Joint} \\ \cline{2-7}
%& EM & F1 & EM & F1 & EM & F1 \\ \hline
%DFGN (paper) & 55.68 & 69.38 & 53.11 & 82.23 & 33.77 & 59.89 \\ \hline
%DFGN + threshold-based \footnote{We finetuned BERT with the released code from DFGN.} & 
%58.27 & 71.90 & 55.79 & 83.57 & 37.12 & 63.04 \\
%DFGN + 2 paragraphs (ours) & 58.74 & 72.53 & 56.61 & 83.57 & 38.08 & 63.87 \\ 
%DFGN + 4 paragraphs (ours)  & 58.80 & 72.67 & 55.48 & 83.34 & 37.07 & 63.63 \\ \hline
%BERT + threshold-based & 57.95 & 71.95 & 53.07 & 82.79 & 35.44 & 62.43 \\
%BERT + 2 paragraphs (ours) & 58.93 & 72.42 & 56.46 & 83.64 & 38.06 & 63.94 \\
%BERT + 4 paragraphs (ours) & 58.70 & 72.67 & 57.65 & 84.86 & 38.19 & 64.24 \\ \hline
%\end{tabular}
%\caption{\label{table:MRC-PS}Results with selected paragraphs on development set in the distractor setting}
%\end{table*}

\section{Case Study}
\begin{figure*}[t!]
\centering
{\includegraphics[width=\linewidth]{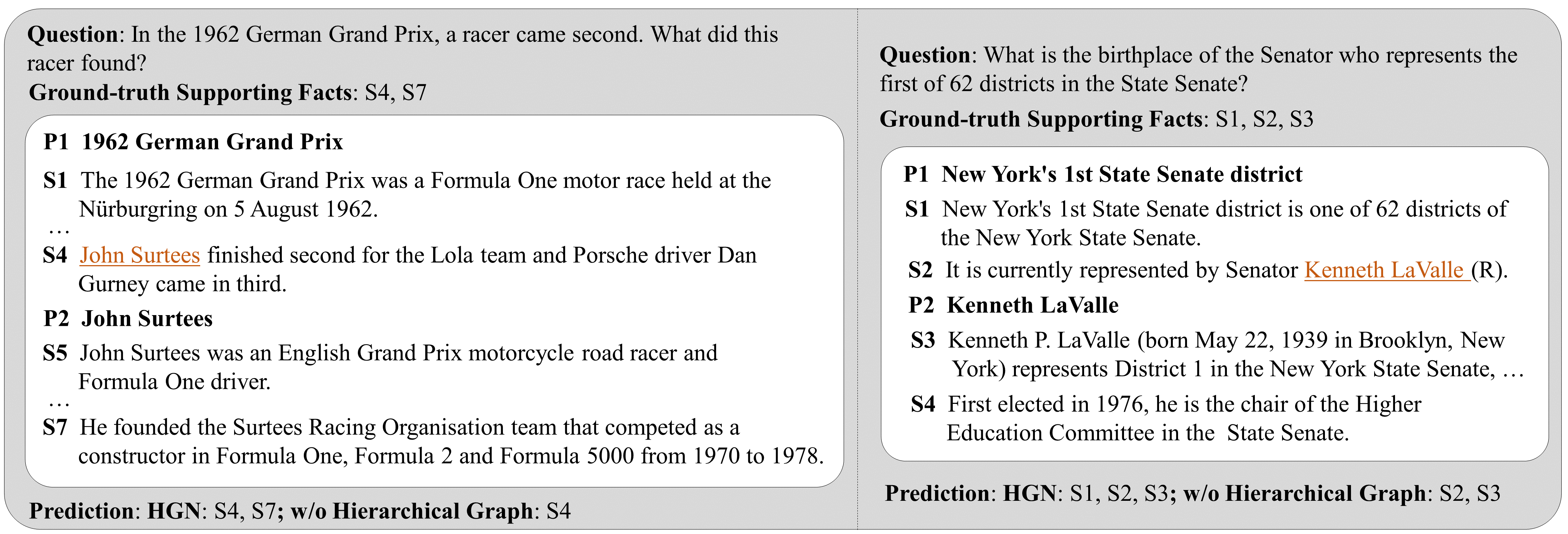}}
\caption{\label{fig:example} Examples of supporting facts prediction in the HotpotQA Distractor setting.}
\label{fig:case_study}
\end{figure*}

We provide two example questions for case study.
To answer the question in Figure \ref{fig:case_study} (left), $Q$ needs to be linked with $P1$. Subsequently, the sentence $S4$ within $P1$ is connected to $P2$ through the hyperlink (\emph{``John Surtees''}) in $S4$. A plain BERT model without using the constructed graph missed $S7$ as additional supporting facts, while our HGN discovers and utilizes both pieces of evidence as the connections among $S4$, $P2$ and $S7$ are explicitly encoded in our hierarchical graph.  

For the question in Figure \ref{fig:case_study} (right), 
the inference chain is $Q \rightarrow P1 \rightarrow S1 \rightarrow S2 \rightarrow P2 \rightarrow S3$. The plain BERT model infers the evidence sentences $S2$ and $S3$ correctly. However, it fails to predict $S1$ as the supporting facts, while HGN succeeds, potentially due to the explicit connections between sentences in the constructed graph.

\begin{table*}[t!]
    \centering
    \begin{tabular}{cp{8cm}}
    \hline
      Category & Sample IDs \\ \hline
      Annotation & 6, 23, 33, 38, 47, 59, 75, 81, 93 \\
      \hline
      Multiple Answers & 1, 4, 8, 10, 11, 16, 19, 24, 26, 28, 29, 32, 39, 40, 42, 50, 53, 56, 60, 63, 67, 68, 71, 72 \\
      \hline
      Discrete Reasoning & 0, 2, 9, 21, 22, 35, 37, 45, 58, 64, 77, 82, 86, 88, 95 \\
      \hline
      Commonsense \& External Knowledge & 7, 15, 20, 36, 69, 70, 73, 76, 78, 83, 84, 85, 87, 91, 92, 96 \\
      \hline
      Multi-hop & 3, 17, 25, 27, 30, 41, 43, 46, 54, 57, 62, 74, 79, 90, 97, 99 \\
      \hline
      MRC & 5, 12, 13, 14, 18, 31, 34, 44, 48, 49, 51, 52, 55, 61, 65, 66, 80, 89, 94, 98 \\ \hline
    \end{tabular}
    \caption{The categories and sample IDs for the 100 examples selected for error analysis. The sample IDs are mapped to the ground-truth IDs in Table~\ref{tab:sample_index}.}
    \label{tab:sample_category}
\end{table*}

\section{Additional Examples for Error Analysis} \label{appendix:additional_examples}

Below, we provide additional examples for error analysis, where ``Q'' denotes question, ``A'' denotes answer provided with dataset and ``P'' denotes the prediction of proposed model. A full list of all the 100 examples is provided in Table~\ref{tab:sample_category} and~\ref{tab:sample_index}. \\

\noindent \textbf{Category: Annotation} \\
\noindent ID: 5ae2e0fd55429928c4239524 \\
Q: What actor was also a president that Richard Darman worked with when they were in office? \\
A: George H. W. Bush \\
P: Ronald Reagan \\
 
\noindent ID: 5ab43b755542991779162c21 \\
Q: What sports club based in Hamburg Germany had a Persian born football player who played for eight seasons? \\
A: Mehdi Mahdavikia \\
P: Hamburger SV \\

\noindent ID: 5a72e28f5542992359bc31ba \\
Q: Which technique did the director at Pzena Investment Management outline? \\
A: outlined by Joel Greenblatt \\
P: Magic formula investing \\

\noindent ID: 5a7e71ab55429949594199bc \\
Q: Perfect Imperfection is a 2016 Chinese romantic drama film starring a south Korean actor best known for his roles in what 2016 television drama? \\
A: Reunited Worlds \\
P: Cinderella and Four Knights \\

\noindent ID: 5a7a18b05542990783324e53 \\
Q: What year was the independent regional brewery founded that currently operates in Hasting's oldest pub? \\
A: since 1864 \\
P: 1698 \\

\noindent \textbf{Category: Multiple Answers} \\
ID: 5a8c9641554299585d9e36f5 \\
Q: Which season of Alias does the English actor, who was born 25 June 1961, appear? \\
A: three \\
P: third season \\

\noindent ID: 5ae6179b5542992663a4f25b \\
Q: Which Hong Kong actor born on 19 August 1946 starred in The Sentimental Swordsman \\
A: Tommy Tam Fu-Wing \\
P: Ti Lung\footnote{Alias of the true answer, Tommy Tam Fu-Wing} \\

\noindent ID: 5abec66b5542997ec76fd360 \\
Q: What do Josef Veltjens and Hermann Goering have in common? \\
A: A veteran World War I fighter pilot ace \\
P: German \\

\noindent ID: 5a85d6d95542996432c570fb \\
Q: What is one element of House dance where the dancer ripples his or her torso back and forth? \\
A: the jack \\
P: Jacking \\

\noindent ID: 5a79c9395542994bb94570a2 \\
Q: Which two occupations does Ronnie Dunn and Annie Lennox have in common? \\
A: singer, songwriter \\
P: singer-songwriter \\

\noindent  \textbf{Category: Discrete Reasoning} \\
\noindent ID: 5a8ec3205542995a26add506 \\
Q: Does Dashboard Confessional have more members than World Party? \\
A: yes \\
P: no \\

\noindent ID: 5abfd83f5542997ec76fd45c \\
Q: Which genus has more species, Quesnelia or Honeysuckle? \\
A: Honeysuckle \\
P: Honeysuckles \\

\noindent ID: 5ac44b47554299194317396c \\
Q: Which became a Cathedral first St Chad's Cathedral, Birmingham or Chelmsford Cathedral? \\
A: Metropolitan Cathedral Church and Basilica of Saint Chad \\
P: St Chad's \\

\noindent ID: 5ac2455e55429951e9e68512 \\
Q: Were both Life magazine and Strictly Slots magazine published monthly in 1998? \\
A: yes \\
P: no \\

\noindent ID: 5a7d26bd554299452d57bb28 \\
Q: Who was born earlier, Johnny Lujack or Jim Kelly? \\
A: Jim Kelly \\
P: John Christopher Lujack \\

\noindent \textbf{Category: Commonsense \& External Knowledge} \\
\noindent ID: 5ac275e755429921a00aaf81 \\
\noindent Q: From what nation is the football player who was named Man of the Match at the 2001 Intercontinental Cup? \\
\noindent A: Ghana \\
\noindent P: Ghanaian \\

\noindent ID: 5ac02d345542992a796decc0 \\
\noindent Q: Where are Abbey Clancy and Peter Crouch from? \\ 
\noindent A: England \\
\noindent P: English \\

\noindent ID: 5ab2beba554299166977408f \\
\noindent Q: Who is the father of the Prince in which William Joseph Weaver is most famous for painting a full length portrait of? \\
\noindent A: George III \\
\noindent P: Queen Victoria \\

\noindent ID: 5a8dab16554299068b959d89 \\
\noindent Q: What type of elevation does Aldgate railway station, Adelaide and Aldgate, South Australia have in common? \\
\noindent A: Hills \\
\noindent P: kilometres \\

\noindent ID: 5a82edae55429966c78a6a9f \\
\noindent Q: Swiss music duo Double released their best known single "The Captain of Her Heart" in what year? \\
\noindent A: 1986 \\
\noindent P: 1985 \\

\noindent \textbf{Category: Multi-hop} \\
\noindent ID: 5a7a46605542994f819ef1ad \\
\noindent Q: What year did Roy Rogers and his third wife star in a film directed by Frank McDonald? \\
\noindent A: 1945 \\
\noindent P: 1946 \\

\noindent ID: 5a84f7255542991dd0999e33 \\
\noindent Q: Which country borders the Central African Republic and is south of Libya and east of Niger? \\
\noindent A: Republic of Chad \\
\noindent P: Sudan \\

\noindent ID: 5a77152355429966f1a36c2e \\
\noindent Q: What was the Roud Folk Song Index of the nursery rhyme inspiring What Are Little Girls Made Of? \\
\noindent A: 821 \\
\noindent P: 326 \\

\noindent ID: 5a7e7c725542991319bc94be \\
\noindent Q: In what year did Farda Amiga win a race at the Saratoga Race course? \\
\noindent A: (foaled February 1, 1999) \\
\noindent P: 1872 \\

\noindent ID: 5ae21ef35542994d89d5b35d \\
\noindent Q: What college teamdid the point guard that led the way for Philedlphia 76ers in the 2017-18 season play basketball in? \\
\noindent A: Washington Huskies \\
\noindent P: University of Kansas \\

\noindent \textbf{Category: MRC} \\
\noindent ID: 5ae5cf625542996de7b71a22 \\
\noindent Q: What sports team included both of the brothers Case McCoy and Colt McCoy during different years? \\
\noindent A: University of Texas Longhorns \\
\noindent P: Washington Redskins \\

\noindent ID: 5a8fa4a5554299458435d6a3 \\
\noindent Q: What is name of the business unit led by Tina Sharkey at a web portal which is originally known as America Online? \\
\noindent A: Sesame Street \\
\noindent P: community programming \\

\noindent ID: 5a8135cc55429903bc27b943 \\
\noindent Q: In the USA, gun powder is used in conjunction with this to start the Boomershot. \\
\noindent A: Anvil firing \\
\noindent P: an explosive fireball \\

\noindent ID: 5a84bb825542991dd0999dbe \\
\noindent Q: Who beacme a star as a comic book character created by Gerry Conway and Bob Oksner? \\
\noindent A: Megalyn Echikunwoke \\
\noindent P: Stephen Amell \\

\noindent ID: 5a75f1a755429976ec32bcb1 \\
\noindent Q: Which actress played a character that dated Mark Brendanawicz? \\
\noindent A: Rashida Jones \\
\noindent P: Amy Poehler \\

\begin{table*}[t!]
    \centering
    \begin{tabular}{lclc}
    ID &  & ID  &  \\ \hline
    0 & 5ac2455e55429951e9e68512 & 1 & 5a8c9641554299585d9e36f5 \\
2 & 5a8ec3205542995a26add506 & 3 & 5a7a46605542994f819ef1ad \\
4 & 5ae6179b5542992663a4f25b & 5 & 5ac3c08a5542995ef918c217 \\
6 & 5ae2e0fd55429928c4239524 & 7 & 5ac275e755429921a00aaf81 \\
8 & 5a7ca98f55429935c91b5288 & 9 & 5a747a9a55429929fddd8444 \\
10 & 5a88696b554299206df2b25b & 11 & 5abec66b5542997ec76fd360 \\
12 & 5ae5cf625542996de7b71a22 & 13 & 5abb729b5542993f40c73af4 \\
14 & 5a85cead5542991dd0999ea9 & 15 & 5ac02d345542992a796decc0 \\
16 & 5a7a88e455429941d65f268c & 17 & 5a84f7255542991dd0999e33 \\
18 & 5a8fa4a5554299458435d6a3 & 19 & 5ae7793c554299540e5a55c2 \\
20 & 5a7755c65542993569682d54 & 21 & 5abfd83f5542997ec76fd45c \\
22 & 5adeb95d5542992fa25da827 & 23 & 5ab43b755542991779162c21 \\
24 & 5a85d6d95542996432c570fb & 25 & 5a8463945542992ef85e23d9 \\
26 & 5ae7d0675542994a481bbdf2 & 27 & 5a82a55955429966c78a6a70 \\
28 & 5ae7313c5542991e8301cbbc & 29 & 5ac44629554299194317395d \\
30 & 5a89d36e554299515336132a & 31 & 5ac2e97d554299657fa290c0 \\
32 & 5a8a764555429930ff3c0de1 & 33 & 5a886211554299206df2b24a \\
34 & 5a8f05b1554299458435d517 & 35 & 5a840e8a5542992ef85e239e \\
36 & 5a7354e35542994cef4bc55b & 37 & 5abc36cc55429959677d6a50 \\
38 & 5a7a18b05542990783324e53 & 39 & 5ab5d27a554299494045f073 \\
40 & 5ac19f405542991316484b5b & 41 & 5a82ebb855429966c78a6a9c \\
42 & 5a72c9e85542991f9a20c595 & 43 & 5ae7739c5542997b22f6a775 \\
44 & 5a84bda45542992a431d1a96 & 45 & 5a7d26bd554299452d57bb28 \\
46 & 5ae21ef35542994d89d5b35d & 47 & 5a753c8c55429916b01642ab \\
48 & 5ac24d725542996366519966 & 49 & 5ae0ec48554299422ee9955a \\
50 & 5a8febb555429916514e73e4 & 51 & 5a7c9d2e55429935c91b5261 \\
52 & 5a8769475542993e715abf2b & 53 & 5abbf519554299114383a0ad \\
54 & 5a735bae55429901807dafef & 55 & 5a7299465542992359bc3131 \\
56 & 5a8b2f2b5542995d1e6f12fa & 57 & 5a77152355429966f1a36c2e \\
58 & 5a87954f5542996e4f308856 & 59 & 5a7e71ab55429949594199bc \\
60 & 5ac531ea5542994611c8b419 & 61 & 5ab72c7d55429928e1fe3830 \\
62 & 5a7e7c725542991319bc94be & 63 & 5ae54c085542992663a4f1c4 \\
64 & 5adc7dbf5542994d58a2f618 & 65 & 5a8fb0be5542997ba9cb32ed \\
66 & 5a8135cc55429903bc27b943 & 67 & 5abcf17655429959677d6b5c \\
68 & 5ab925fd554299131ca42281 & 69 & 5ab2beba554299166977408f \\
70 & 5a8dab16554299068b959d89 & 71 & 5ac38ce255429939154137c2 \\
72 & 5a79c9395542994bb94570a2 & 73 & 5ab946d7554299743d22eaaf \\
74 & 5a73d33e5542992d56e7e3a9 & 75 & 5a72e28f5542992359bc31ba \\
76 & 5ab1d983554299340b52540a & 77 & 5a7cb9b95542990527d55515 \\
78 & 5a773d8955429966f1a36cc4 & 79 & 5a7780e855429949eeb29e9f \\
80 & 5a84bb825542991dd0999dbe & 81 & 5a7698c2554299373536010d \\
82 & 5ae1847e55429920d52343ee & 83 & 5a7199725542994082a3e88f \\
84 & 5abf11d45542997719eab660 & 85 & 5ae52cb955429908b6326540 \\
86 & 5ac44b47554299194317396c & 87 & 5a7d61775542991319bc93b9 \\
88 & 5ae0536755429924de1b70a6 & 89 & 5a75f1a755429976ec32bcb1 \\
90 & 5adbc8e25542996e68525230 & 91 & 5a72ac8a5542992359bc3164 \\
92 & 5adc6ded55429947ff17395d & 93 & 5a7b971255429927d897bff3 \\
94 & 5ae34a225542992e3233c370 & 95 & 5ac2cdaa554299657fa29070 \\
96 & 5a82edae55429966c78a6a9f & 97 & 5a8a3a355542996c9b8d5e5e \\
98 & 5adcc90c5542990d50227d1b & 99 & 5a79c9c05542994bb94570a5 \\
\hline
    \end{tabular}
    \caption{The full index list of the 100 samples selected for error analysis.}
    \label{tab:sample_index}
\end{table*}

\end{document}